# Large Language Model Predicts *Above Normal* All India Summer Monsoon Rainfall in 2024

Ujjawal Sharma[1], Madhav Biyani[2], Akhil Dev Suresh[3], Debi Prasad Bhuyan[3], Saroj Kanta Mishra[3*], Tanmoy Chakraborty[2,4 *]

[1]Department of Computer Science and Engineering, IIT Delhi, India
[2]Department of Electrical Engineering, IIT Delhi, India
[3]Centre for Atmospheric Sciences, IIT Delhi, India
[4]Yardi School of Artificial Intelligence, IIT Delhi, India

*Corresponding Author:* Tanmoy Chakraborty (tanchak@iitd.ac.in), Saroj Kanta Mishra (skm@iitd.ac.in)

## Abstract

Reliable prediction of the All India Summer Monsoon Rainfall (AISMR) is pivotal for informed policymaking for the country, impacting the lives of billions of people. However, accurate simulation of AISMR has been a persistent challenge due to the complex interplay of various muti-scale factors and the inherent variability of the monsoon system. This research focuses on adapting and fine-tuning the latest LLM model, PatchTST, to accurately predict AISMR with a lead time of three months. The fine-tuned PatchTST model, trained with historical AISMR data, the Niño3.4 index, and categorical Indian Ocean Dipole values, outperforms several popular neural network (NN) models and statistical models. This fine-tuned LLM model exhibits an exceptionally low RMSE percentage of 0.07% and a Spearman correlation of 0.976. This is particularly impressive, since it is nearly 80% more accurate than the best-performing NN models (here, CNN). The model predicts an above-normal monsoon for the year 2024, with an accumulated rainfall of 921.6 mm in the month of June-September for the entire country.

# Introduction

The South Asian Summer Monsoon (SASM) is considered a lifeline for the Indian subcontinent[1], contributing approximately 75% of India's annual rainfall and supporting the water demands of billions of people. It plays a crucial role in multiple sectors, including agricultural planning, hydro-power generation, industrial expansion, etc. This makes accurate and unbiased seasonal monsoon rainfall prediction essential for effective planning across various sectors in India[2]. Although the contribution of agriculture to the country's GDP has shrunk from approximately 35% to 15% over the past three decades, a large chunk of agricultural activities are still strongly tied to the precipitation cycle during the summer monsoon season[3]. Therefore, enhancing the precision of monsoon forecasts can significantly improve resource management and bolster the region's economic stability.

The Indian Summer Monsoon Rainfall (ISMR) exhibits a large inter-annual variability of ~10%, with large spatial variability in the precipitation distributions from year-to-year, due to its linkage with many local factors and remote coupled ocean-atmospheric processes, influencing the ISMR largely in a wide range of timescales, from intra-seasonal to decadal[4]. Apart from these natural variabilities in ISMR, a vast body of literature suggests that the warming climate, anthropogenic emissions, and changes in land-use-land-cover contribute to increased monsoon variability[5,6]. These changes make the ISMR highly non-linear and extremely challenging to understand and make accurate predictions. The resulting variability in ISMR impacts billions of people and their livelihoods, with profound implications for the country's GDP[7].

Substantial advancement has been made in the long-range prediction of ISMR by the Indian Meteorological Department (IMD), starting from a simple statistical model with a small number of predictors to the involvement of both statistical models with many potential predictors of the summer monsoon and the ensemble mean of state-of-the-art (SOTA) coupled models in recent days[8]. Many studies have highlighted the introduction of coupled models for ISMR prediction, resulting in substantial improvement in prediction skills[8]. This approach is also extensively used by the leading seasonal prediction centres[9,10]. The prediction skill of these physical models largely depends on the initial conditions, model physics, slowly varying boundary conditions, ocean-atmosphere coupling, and the loosely constrained model parameters involved in the parameterization processes to represent the sub-grid scale phenomenon[11,12]. These processes are computationally resource-intensive, and the differential equations involved in modeling the dynamics of earth systems are often completely unsolvable and are more than often approximated. As a result, these models are unable to serve the purpose of a reliable seasonal prediction, with predictions contradicting the ground truth many a times, thereby limiting its effectiveness.

In recent decades, the advent of emerging technologies like artificial intelligence (AI) and machine learning (ML) have emerged as effective tools in various fields such as energy, weather, transportation, healthcare, finance, agriculture, etc. One of the important applications in the field of atmospheric sciences is time-series forecasting using AI/ML[13-15], as many existing methods use machine learning and deep learning techniques and have shown promising results. In many instances, these models outperform the conventional physical models along with an advantage of minimal workforce and computational resources[13].

These machine learning models have demonstrated exemplary performance in capturing the complex non-linear relationships between the input features and the target output, thereby allowing them to capture complex real-world dynamic systems, e.g., weather forecasting, energy distribution, traffic flow analysis, and financial market modeling. Among the presently available deep learning models, the pre-trained Large Language Models (LLMs) have become increasingly popular in recent years because of their skill in time-series forecasting with limited training datasets as input owing to their large-scale pretraining over mountains of data[16-18]. Among the presently available LLMs, the channel-independent patch time series Transformer (PatchTST) (see the methods for more details) has shown excellent performance in long-term time series forecasting and outperforms many SOTA LLM-based models[19].

The capabilities of LLMs in seasonal forecasting of key atmospheric variables remain largely unexplored and need detailed investigations. Our study examined the skill of many popular ML models that are extensively utilized for timeseries forecasting, including three statistical models (Linear Regression, XG Boost, Support Vector Regression (SVR)), two neural network (NN) models (Long Short-Term Memory (LSTM), Convolution Neural Network (CNN)), along with one LLM model (PatchTST) in forecasting the All India Summer Monsoon Rainfall (AISMR). It is noteworthy that our study represents one of the first attempts to explore the implications of LLMs in AISMR forecasting.

# Results

## a. Model performance

The PatchTST model demonstrates exceptional performance, characterized by a very low RMSE percentage and a high Spearman correlation value (Fig 1). When compared with baseline models, all versions of the PatchTST model exhibit significantly higher Spearman correlation values, indicating their proficiency in accurately capturing the rank of the observations. This translates to the model's ability to capture the variability in rainfall effectively. Furthermore, these models exhibit a low RMSE percentage error, with the best-performing model achieving an RMSE percentage as low as 0.07%.

Among the four versions of the PatchTST, the model fine-tuned with the AISMR+Niño3.4+IOD (named as PatchTST_AISMR+Niño3.4+IOD) dataset emerges as the best-performing model, boasting a Spearman correlation of 0.967 and an RMSE percentage error of 0.07%. This is followed by the model fine-tuned on the AISMR+IOD (PatchTST_AISMR+IOD) dataset, with a Spearman correlation value of 0.889 and an RMSE percentage error of 0.25%. The model trained with the AISMR (PatchTST_AISMR) data has an RMSE percentage error of 0.27% and a Spearman correlation value of 0.816. Finally, the model trained with the AISMR+ Niño3.4 (PatchTST_AISMR+ Niño3.4) dataset has the largest RMSE percentage error among all the different versions of the PatchTST model, with an error of 0.79% models and a lower Spearman correlation of 0.786. However, even this relatively poorer-performing PatchTST model is comparable to the best-performing baseline models, underscoring the superior performance of the PatchTST approach. Moreover, the PatchTST model demonstrates the capability to accurately predict the AISMR while requiring significantly less computational resources. This is a salient feature of the model, especially since physical models require considerable computational resources and a skilled workforce to make predictions. Comparing the forecasts from the best-performing PatchTST model, PatchTST_AISMR+Niño3.4+IOD, with observed cumulative rainfall, the model predicts the rainfall with high accuracy (Fig. 2) for the test period (2011-2023). The deviation of the model predictions with the observed rainfall is less than 1% during the test period and agrees well with the observed interannual variations of AISMR.

The PatchTST model's outstanding performance in predicting cumulative rainfall could be attributed to the model's unique feature of patching and channel segmentation, enabling the model to retain local semantic information essential for AISMR prediction. This underscores the model's potential as a reliable tool for rainfall forecasting. Its superior performance compared to baseline models, coupled with its resource efficiency make it a promising tool for supplementing conventional physical models that require extensive computational resources.

**b. Rainfall prediction for 2024**

Our model predicts an "above normal" (based on IMD tercile category[13]) monsoon for the year 2024 with a cumulative rainfall of 921.6 mm. The model can make these predictions with a lead time of three months, which is early enough for adopting appropriate preparedness measures to tackle any unforeseen situations. The best-performing model utilizes the historical AISMR data along with the Niño3.4 index and IOD index and is able to understand the complex interaction among them, resulting in accurate prediction of AISMR. This demonstrates the ability and applicability of powerful data-driven algorithms to exploit the available data for a multitude of public-relevant applications. A similar approach could also be employed for region-specific predictions of summer monsoon rainfall; as a matter of fact, any climatic prediction provided an appropriate dataset.

**c. Impact of input features on model performance**

The PatchTST_AISMR+Niño3.4 model exhibited the lowest performance from the rest of the PatchTST model versions (see Fig. 1), which was unexpected given the extensive literature emphasizing the potential teleconnection between AISMR and the Niño3.4 index[20-22]. Based on these studies, we anticipated that the PatchTST_AISMR+Niño3.4 model would outperform the PatchTST_AISMR model. Interestingly, the model with the best performance utilized Niño3.4 values in conjunction with IOD and AISMR values for its predictions. This suggests that while there exists a relationship between Niño3.4 and monsoon rainfall, this connection alone is insufficient for accurately predicting AISMR. One possible explanation for this finding is the evolving teleconnection between monsoon rainfall and the El Niño-Southern Oscillation, showing a decreased teleconnection strength since the last four decades[21]. The changing nature of this teleconnection could contribute to the reduced effectiveness of using AISMR and Niño3.4 data alone for model predictions.

This highlights the importance of carefully curating the input features for the model to utilize the maximum potential of the models. The choice of input features directly influences the model's ability to capture complex relationships and patterns within the data, ultimately determining the accuracy and reliability of the model's predictions. Therefore, careful consideration of the input features is paramount in ensuring that the model can make the most of the available information and deliver optimal performance.

## Conclusion

The data-driven LLM, PatchTST model successfully simulated the seasonal forecast of AISMR for all years during the test period with minimal biases, showcasing a new and efficient approach to the resource-intensive AISMR forecast. This developed model could complement existing physical models to enhance the seasonal forecast skill for AISMR prediction. The PatchTST model predicts an "above normal" monsoon rainfall season for 2024, achieving this prediction with a lead time of three months. This early prediction would significantly benefit agricultural activities and water resource management, thereby supporting the country's overall economy. However, providing a point forecast for AISMR is not sufficient to address the need for actionable seasonal forecasts. It necessitates granular-scale seasonal forecasts across the entire Indian subcontinent. This requires adopting data-driven models for individual grid scales to provide socially relevant seasonal monsoon rainfall information. Given that the IMD provides gridded data with a resolution of 0.25 degrees, it is practically feasible to perform regional forecasts at a granular scale by fine-tuning the model parameters for specific regions.

## Methods

The models are trained on four different datasets. The first dataset consists of AISMR daily data for the months of JJAS. The second dataset includes the AISMR daily data and monthly Niño3.4 index values from the month of May of the previous year (T) till the month of May of the prediction year (T+1). The third dataset is composed of AISMR daily values for the month of JJAS combined with IOD index values of the prediction year (T+1). The final dataset incorporates the AISMR daily data, 13 months of Niño3.4 index values, and the IOD index values. The IOD index is categorically classified as positive (+1), negative (-1), or neutral (0) in the input dataset.

### Baseline models

The baseline models are based on the framework outlined by Narang et al. (2023). These baseline models comprise of three statistical models and two NN models.

a. Statistical models

    The statistical model includes Linear Regression, XGBoost, and Support Vector Regression (SVR)these models are trained on four different datasets with five varying lookback periods, resulting in a total of 20 different input datasets. The model's performance is measured and compared against the PatchTST model.

b. Neural Network models

    There are two NN-based models used as baselines in this study, namely LSTM and CNN-based models. Similar to the statistical models, these models are trained on the same 20 different input datasets. The model's performance for each input dataset is measured separately and compared.

**Transformers and PatchTST**

A transformer is a deep learning architecture that employs a multiple-head attention mechanism. The input is divided into tokens in this architecture, and those tokens are subsequently transformed into vectors by using a word embedding table. At each layer, these tokens are contextualized within a specified window, allowing them to interact with each other through a parallel multi-head attention process. This process emphasizes important tokens while less significant ones are downplayed. One key advantage of transformers is the absence of recurrent units, which enables faster training compared to earlier recurrent neural networks (RNNs) like long-short term memory (LSTM). The transformer architecture has also facilitated the development of pre-trained models like generative pre-trained transformers (GPTs) and BERT (Bidirectional Encoder Representations from Transformers). The PatchTST model architecture[18] (the model discussed in this paper), leverages the power of Transformers to process univariate time series data. The model operates by transforming the input data through a series of processing steps, including patching, projection, and Transformer-based encoding (Fig. 3). The model vectorizes individual time series into patches of a specified size and encodes these sequences using a Transformer. This process facilitates the extraction of meaningful features and the generation of forecasts.

The model consists of two key components:

1. Segmentation into Patches: The time series is divided into patches, which are then used as input tokens for the Transformer. This approach helps retain local semantic information and reduces the computational complexity associated with processing long sequences.

2. Channel Independence: Each channel in the model corresponds to a single univariate time series. The model shares embedding and Transformer weights across all channels, making it a global univariate model. This design enhances the model's ability to generalize across different time series.

The patching mechanism offers several advantages:

1. Retention of Local Information: The model preserves important local patterns within the time series by processing patches.
2. Reduced Computational Load: The segmentation into patches reduces the size of the input sequence, thereby decreasing the computational and memory requirements of the model.
3. Enhanced Historical Context: The model can consider longer historical sequences by adjusting the patch and context lengths.

**Model working and fine-tuning**

The PatchTST model processes input data by segmenting time series into patches, each representing a window of consecutive data points. The input features are first scaled using the Standard Scaler to normalize the data. The normalized data is then organized into sequences, where each sequence consists of a fixed number of consecutive data points defined by a window size (in this case, 30). These sequences are used as input tokens for the Transformer. The model architecture includes an LSTM-based encoder with a specified number of layers and hidden dimensions. Each input patch is processed independently, with shared weights across different time series channels, thus maintaining channel independence. This design allows the model to effectively capture local and global temporal dependencies.

Fine-tuning of the model involves adjusting several hyperparameters through grid search. Key parameters such as the number of LSTM layers, hidden dimensions, dropout rate, and learning rate are optimized to improve model performance. For instance, the final model configuration includes 128 hidden units, 3 LSTM layers, and a learning rate of 0.001 with a batch size of 64. These choices were made to balance the model's capacity and generalization ability, ensuring that it captures complex patterns in the data without overfitting.

To enhance training stability and prevent gradient explosion, gradient clipping is applied. Additionally, early stopping criteria are implemented to halt training if no significant improvement in the loss is observed over multiple epochs, thereby preventing overfitting and ensuring optimal model performance.

For model training, 110 years of data (1901-2010) were utilized, while the most recent 13 years (2011-2023) were reserved for testing the model's predictive capabilities.

**Model evaluation**

The outputs of the trained models were evaluated using different metrics, these include –

RMSE Percentage: The root mean squared error (RMSE) percentage represents the difference between the actual and predicted values as root mean squared error percentages, facilitating direct comparison between models. The better performing models generally has the lowest RMSE percentage values.

$$RMSE(\%) = \sqrt{\frac{100}{n} \sum_{i=1}^{n} \left(\frac{Obs_i - Pred_i}{Obs_i}\right)^2}$$

Spearman Correlation: Spearman Correlation: Spearman correlation represents the statistical measure of the strength of the relation between a ranked pair of variables. The Spearman correlation varies between −1 and +1, where positive correlation implies that if one variable increases, the other also increases and vice-versa. Similarly, a negative correlation implies if one variable decreases, the other variable increases and vice-versa. A correlation between the actual and predicted values closer to 1 signifies better model performance.

$$Spearman\ correlation = 1 - \frac{6 \sum_{i=1}^{n} d_i^2}{n^3 - n}$$

## Data availability

The primary dataset used in this study includes daily rainfall data for the months of June to September from the Indian Meteorological Department (IMD)[23]. The record spans 123 years, from 1901 to 2023. Additionally, monthly values of the Niño3.4 index (Niño3.4) and the Indian Ocean Dipole (IOD) index are used as predictors in this study due to their strong teleconnection with ISMR[13]. The Niño3.4 values are archived and maintained by the National Oceanic and Atmospheric Administration (https://psl.noaa.gov/gcos_wgsp/Timeseries/Data/nino34.long.anom.data), and the IOD index values are maintained by the Japan Meteorological Agency (https://psl.noaa.gov/gcos_wgsp/Timeseries/Data/nino34.long.anom.data).

For the prediction of 2024 the multi-model mean forecasts of Niño3.4 index provided by the Columbia Climate School (https://iri.columbia.edu/our-expertise/climate/forecasts/enso/current/) and monthly IOD forecasts provided by the Bureau of Meteorology Australia (http://www.bom.gov.au/climate/enso/#tabs=Indian-Ocean) have been used.

## Code availability

All codes used in this study are available from the corresponding authors upon request.

## Acknowledgements

This work is partly supported by the DST Centre of Excellence for Climate Information at the Indian Institute of Technology Delhi. In addition, we acknowledge the generous funding support from Tower Research Capital Markets towards using machine learning for social good.

## Authors contributions

SKM and TC conceived the study and designed the work. All the authors performed the analyses and contributed to writing the manuscript.

## Competing interests

The authors declare no competing interests.

# List of figures

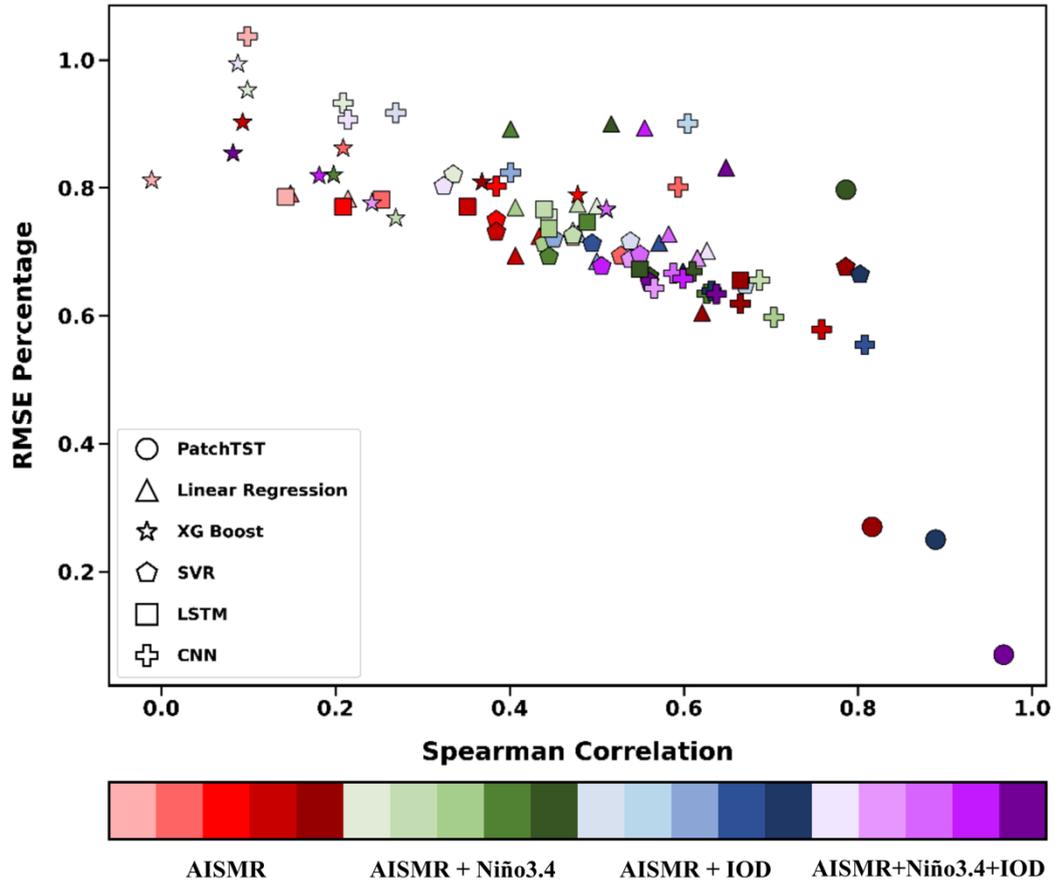

**Fig 1.** Performance of PatchTST and baseline models. The X-axis represents the Spearman correlation, while the Y-axis denotes the RMSE percentage. Marker shapes indicate various models, and shading highlights the different datasets used. Note that the PatchTST model does not have a lookback period in the dataset, the distinct colors represent the different dataset used as inputs irrespective of the lookback period.

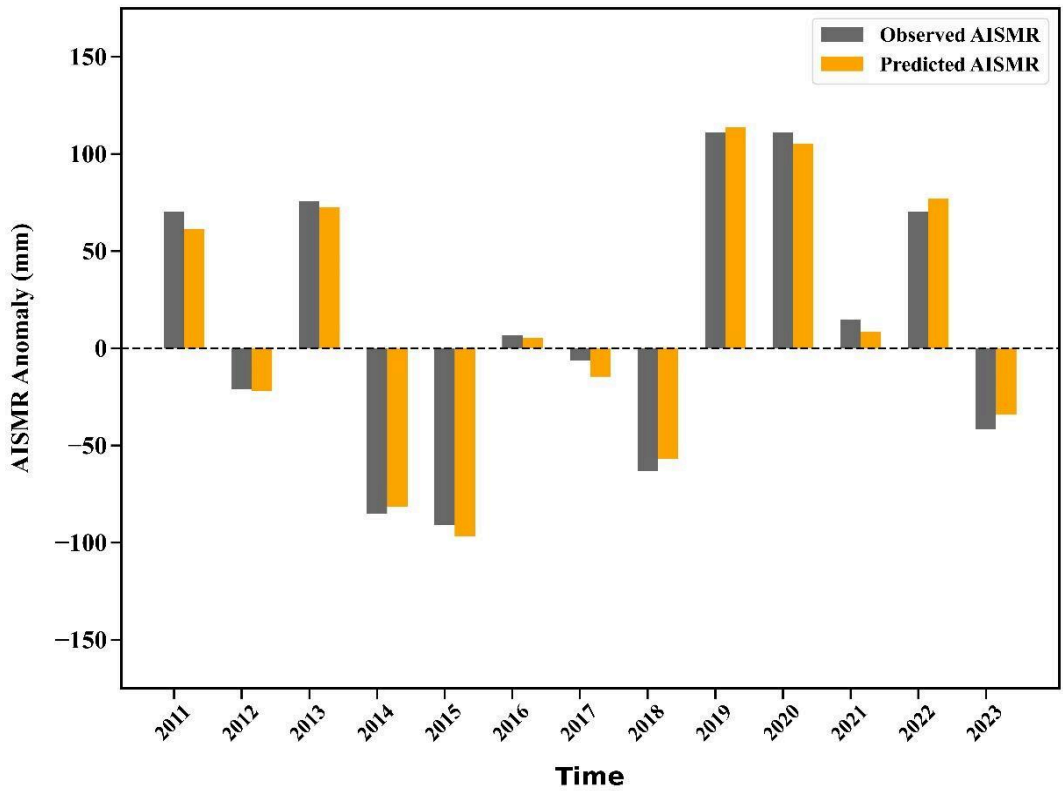

**Fig 2.** All India Summer Monsoon Rainfall Anomaly for the test period (2011-2023). The accumulated rainfall for the months of JJAS is subtracted from the long-term mean rainfall value for both observations and model predictions.

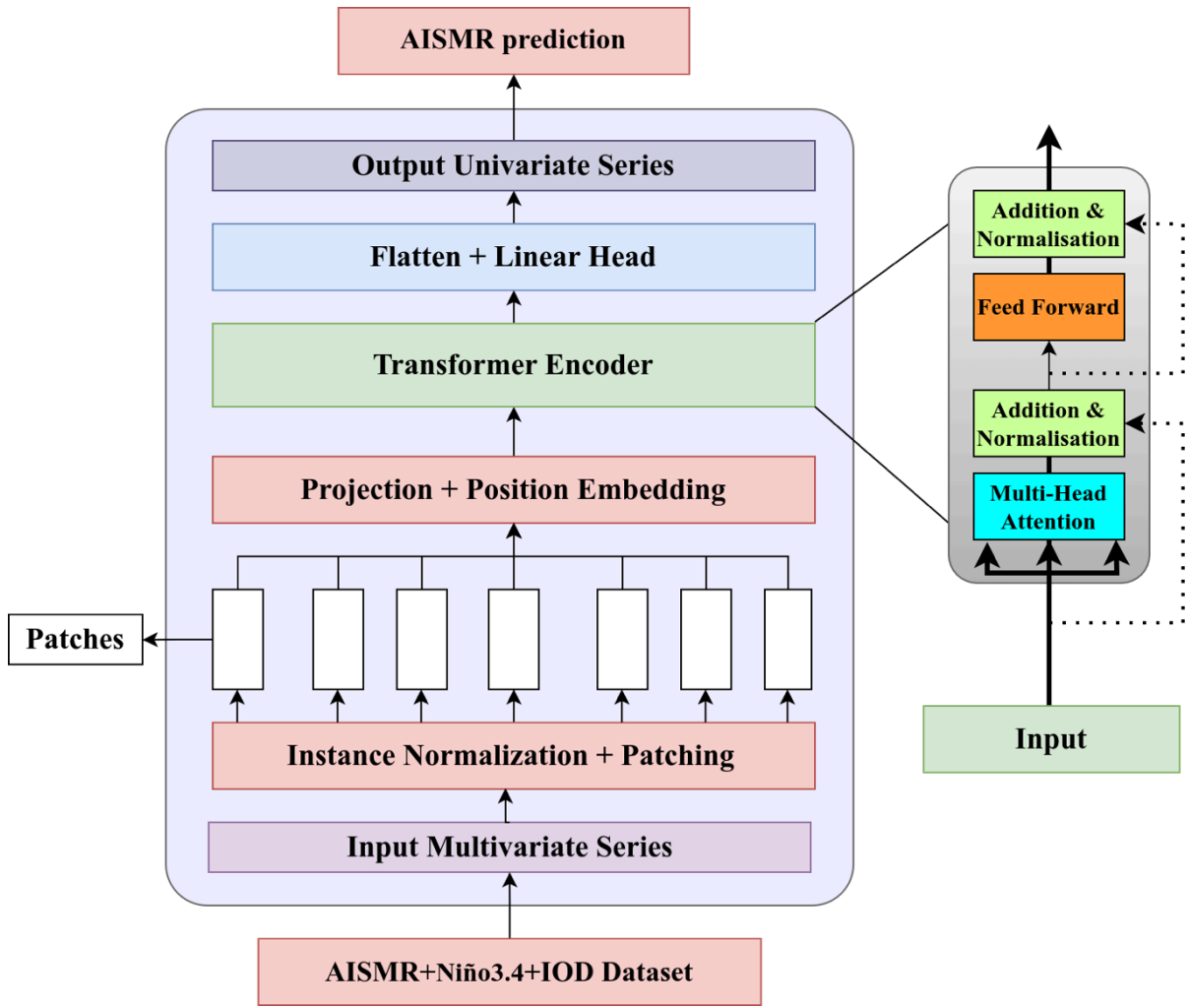

**Fig 3.** Schematic of the PatchTST (Nie et al. 2022) model employed in this study.